\def\year{2017}
\documentclass[letterpaper]{article}
\usepackage{aaai}
\usepackage{mathpartir}
\usepackage{times}
\usepackage{helvet}
\usepackage{courier}
\usepackage{amsmath}
\usepackage{amssymb}
\usepackage{graphicx}
\usepackage{color}
\usepackage{comment}
\usepackage{lettrine} 
\usepackage[font=small,skip=8pt]{caption}
\usepackage{float}
\makeatletter
\newcommand{\verbatimfont}[1]{\renewcommand{\verbatim@font}{\ttfamily#1}}
\makeatother
\title{Neural Semantic Parsing Natural Language into SQL}
\author{Ruiyang Xu, Ayush Singh\\
\{xu.r, singh.ay\}@husky.neu.edu\\
CS6120 NLP Fall 2017, Northeastern University
}

\begin{document}
%
\maketitle

\section{Introduction}
Natural interface to database (NLIDB) has been researched a lot during the past decades. In the core of NLIDB \cite{popetheorynli} is a semantic parser used to convert natural language into SQL. For instance, given a database schema information:
\begin{figure}[H]
\begin{minipage}{0.48\textwidth}
\centering
\includegraphics[width=\textwidth]{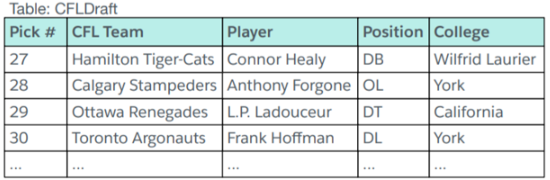}
\end{minipage}
\end{figure}
\begin{figure}[H]
\begin{minipage}{0.3\textwidth}
\centering
\vspace*{-0.7cm}
\includegraphics[width=\textwidth]{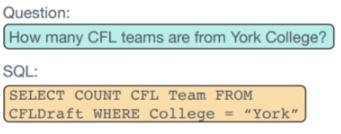}
\end{minipage}
\end{figure}
\vspace*{-0.3cm}
Solutions from traditional NLP methodology focuses on grammar rule pattern learning \cite{wang2017synthesizing} and paring via intermediate logic forms \cite{katelearningtransn2f}. Although those methods give an acceptable performance on certain specific database and parsing tasks, they are hard to generalize and scale. On the other hand, recent progress in neural deep learning seems to provide a promising direction towards building a general NLIDB system. Unlike the traditional approach, those neural methodologies treat the parsing problem as a sequence-to-sequence learning problem. In this project, we experimented on several sequence-to-sequence learning models and evaluate their performance on general database parsing task.  

\section{Related Work}
This project was motivated by a recent research of Seq2SQL from Salesforce \cite{salesforce}. In their paper, they have mentioned a reinforcement learning enhanced attention pointer network model. The accuracy of their model is 59\%. However, as claimed in their paper, the reinforcement learning only helps to increase the accuracy by only 2\%. A previous work \cite{ptrnet} which only use pointer network but has a tree-like decoder instead of a sequence decoder can already achieve an accuracy of 36\%. And we have evidence to think Seq2SQL has also used a tree decoder. 

The pointer network \cite{ptrnet} has gained a lot of attention in recent years. It is a location-based attention mechanism which tries to put attention on the locations in the input sequence. It tries to compose an output from the components in the input via predicting pointers to the index in the input. This structure has even been proved to have the power to solve combinatorial optimization problems within a certain size. As mentioned by the author, this attention mechanism was a modification of a contention based attention mechanism \cite{attention} which put attention on the contents of the input sequence. Unlike pointer network, it tries to extract input-output correlation information and use that information to assist decoder inference and it will still use output vocabulary to compose its output. Underlying those attention mechanisms is a vanilla sequence to sequence model \cite{seq2seq} or encoder-decoder model \cite{rnnencdec}

Considering the complexity of tree decoders and time limitation of this project, we implemented a sequence decoder instead. And since Salesforce decided to keep their implementation close sourced, we have to implement our own pointer decoder (unfortunately, Tensorflow hasn't incorporated pointer network for the time being). And we notice that research on pointer network based sequence to sequence model on semantic parsing has rarely been done by other researchers. Hence we built the whole end to end system on our own with Tensorflow. We also build two other models to compare the performance among those sequence to sequence models, namely "vanilla seq2seq", "seq2seq with contention-based attention mechanism" and "seq2seq with pointer network". To mitigate the negative effect caused by using padding symbols with attention mechanism, we introduced an emphasizing strategy (will mention later). 

SQLizer \cite{yaghmazadehsqlizer} was our primary inspiration by using semantic parsing from the English description into an intermediate logic form, a query sketch (which is essentially a relational algebra with holes). This sketching technique allows them to utilize technologies from program synthesis community (a type theory guided sketch completion). Yet the authors haven't further discussed the shortcoming and regression performance of SQLizer on medium to large datasets with complex queries rather than tested it on a handcrafted small dataset. Therefore the scalability of their method is still questionable.

\section{Dataset}
We are using the WikiSQL dataset \cite{salesforce}.
\begin{enumerate}
\item A large-scale dataset (80,654 records with 24,241 schemas) suitable for effectively training neural networks.
\item Crowd-sourced to collect the natural language questions created by human beings, that helps overcome overfitting to template-synthesized descriptions.
\item  The task synthesizes the SQL query based only on the natural language and the table schema
without relying on the table’s content, that helps mitigate scalability as well as privacy issue that alternative approaches \cite{yaghmazadehsqlizer} may suffer when being applied to realistic application scenarios where large-scale and sensitive user data is involved.
\item The data is split so that the training (70\%), dev (10\%), and test (20\%) set do not share tables which helps evaluate an approach’s capability to generalize to an unseen schema. 
\end{enumerate}
Note that the WikiSQL task considers synthesizing a SQL query with respect to only one table. Thus, in an output SQL query, only the SELECT clause and the WHERE clause need to be predicted, and the FROM clause can be omitted.

\section{Methodology}
\subsection{Preprocessing}
Since the original dataset is serialized for storage. We have found a way to deserialize the dataset and use them as our training, dev and test data.
\begin{figure}[H]
\begin{minipage}{0.5\textwidth}
\centering
\hspace*{-0.6cm}
\includegraphics[width=\textwidth]{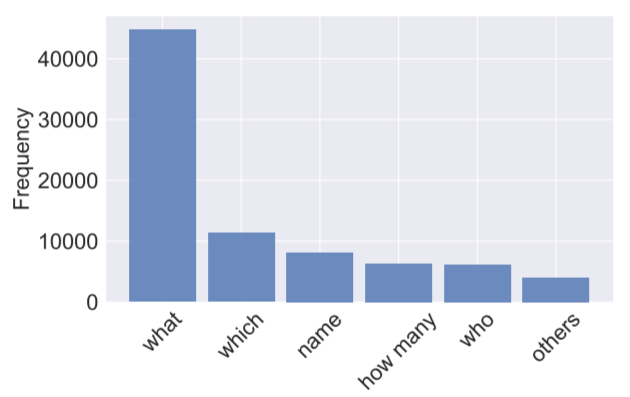}
\caption{Distribution of question types in WikiSQL}
\end{minipage}
\end{figure}
We first removed all punctuations from the input data but then realized it disturbs names of the column names which made evaluation tricky so we only removed ? from the end of sentences. Further, we replaced column names with multiple words with single words via adding hyphens and also did some alignment between column names appeared in questions and tables, so that the neural network won't treat them as two different objects. The final vocabulary size for questions and SQL respectively are 57,625 and 44,554.

\begin{figure}[H]
\begin{minipage}{0.49\textwidth}
\centering
\hspace*{-0.3cm}
\includegraphics[width=1\textwidth]{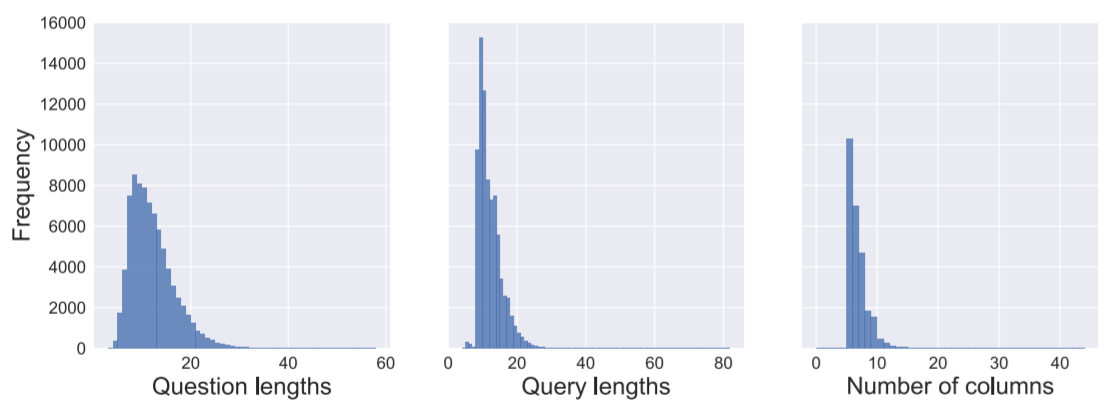}
\caption{Distribution of question, query, table in WikiSQL}
\end{minipage}
\end{figure}

For the vanilla seq2seq model, the preprocessing above is good enough and no need to do further process. However, we have to augment the input sequence in order to provide more contextual information to the attention mechanism. \\\\
\textbf{Original}: What is the description of a ch-47d-chinook \\
\textbf{Column}: Aircraft/VARCHAR Description/VARCHAR Max Gross Weight/INT Total disk area/INT Max disk Loading/INT \\
\textbf{Seq2SQL Augmented}: \textless col\textgreater;Aircraft Description Max-Gross-Weight Total-disk-area Max-disk-Loading;\textless sql\textgreater;SELECT FROM WHERE COUNT MIN MAX AVG SUM AND = ;\textless question\textgreater; What is the description of a ch-47d-chinook \\
\textbf{Augmented}: SELECT FROM WHERE COUNT MIN MAX AVG SUM AND = ; Aircraft Description Max-Gross-Weight Total-disk-area Max-disk-Loading ; What is the description of a ch-47d-chinook

The Sentinel (;) tokens are only for descriptive purposes to demarcate boundaries in our data unlike Seq2SQL and are removed in the original implementation. The first part comprises of all column names followed by SQL keywords and finally the natural language question. 
Bringing SQL keywords in the front helps the Pointer network easily point to SQL positions which are bound to change in case of Seq2SQL augmented data due to variable column names for different schemas. 

\subsection{Vanilla Seq2Seq Model}
Our vanilla seq2seq model has a sequential encoder and decoder. The encoder is a 2 layers dynamic LSTM neural network. The hidden state of the last layer will be used as the initial state for the decoder. The decoder is also 2 layers dynamic LSTM neural network with a dense projection layer as final prediction output. We applied the word to vector embeddings on both encoder and decoder inputs with embedding size 300. We used weighted cross-entropy as our loss function. We also applied Adam optimizer with gradient clipping as the final trainer (learning rate was set to 0.01).  

\subsection{Reversed Seq2Seq Model}
\cite{seq2seq} found that if we only reverse the source sequence and let target be intact, it reduces the distance between translation words and hence increase memory ability on long sentences while reducing computation time.

In our experiments, we did not find reversing technique to be any superior, and would like to point that it is case specific.

\subsection{Bidirectional Seq2Seq Model}
Traditional LSTM are unidirectional which means they take only the word appearing after them into context. Adding another layer in parallel that takes previous word into context and merging/concatenating/averaging both layers allows the model to take both neighbor words into context and is found to be outperforming unidirectional LSTMs.

\subsection{Attention Seq2Seq Model}
We modified our vanilla seq2seq model to add in Bahdanau attention mechanism \cite{attention} with Bidirectional LSTMs. Except for using augmented dataset, other hyperparameters kept intact.

\subsection{Pointer Attention Seq2Seq Model}
Since Tensorflow didn't provide attention wrapper for pointer network, we have to hack the Tensorflow kernel code to modify the existing Bahdanau Attention Mechanism into pointer network. This modification effected our previous architecture tremendously. First, after a long time of trial and error, we noticed that it is impossible to implement pointer network with dynamic LSTM, so we have to change our design to static LSTM with fixed cell size (which means we have to pad all our input data to certain fixed length). Another trouble caused by this modification attempt is that the encoder and decoder will have the same cell size! This problem is more serious than the previous one because, averagely, output sequences are much shorter than input sequences. We noticed that padding will affect the probability distribution dramatically once using attention mechanism. To mitigate this effect, instead of using single padding symbols, we used the replication of input sequence. Namely, we will replicate an input sequence until it achieves the specified cell size. We called this strategy \textit{emphasizing}.

\section{Experiments}
\subsection{Dataset and Metric}
We trained our models on a training set of 60000 data and tested them on a test set of 9145 data. However, we later noticed that it is really difficult for a sequential structural LSTM to memorize and learn a long sequence; the exact match accuracy is insignificant for all three models. So we have to change our metric to component-wise match, where we count in an output sentence how many positions have been correctly predicted in case of pointers and bag of words to calculate average accuracy over the whole sequence.

\begin{table}[!h]
\label{Models Accuracy Comparison}
\begin{center}
\begin{tabular}{| l | c | c |}
\hline
\textbf{Seq2Seq Models} & \multicolumn{2}{ c |}{\textbf{Accuracy \%}}  \\ 
\cline{2-3}
& \textbf{Train} & \textbf{Validation} \\
\hline

Vanilla & 74.43 & 50.66 \\ \hline
Reversed & 79.37 & 49.96 \\ \hline
Bidirectional & \textbf{81.28}  & \textbf{56.29} \\ \hline
Attention & 100 & 48.27 \\ \hline
Pointer Attention & 17.61 & 14.12  \\ \hline

\end{tabular}
\caption{Accuracy and loss for pointer network}
\end{center}
\end{table}

\begin{figure}[H]
\centering
\includegraphics[width=0.5\textwidth]{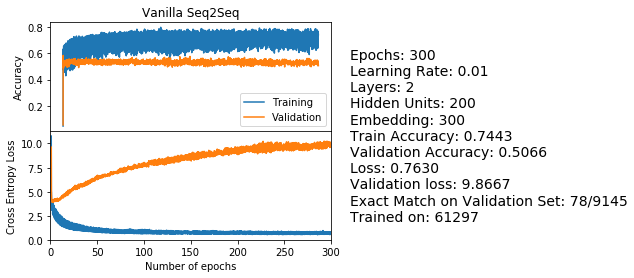}
\vspace*{0.6cm}
\includegraphics[width=0.5\textwidth]{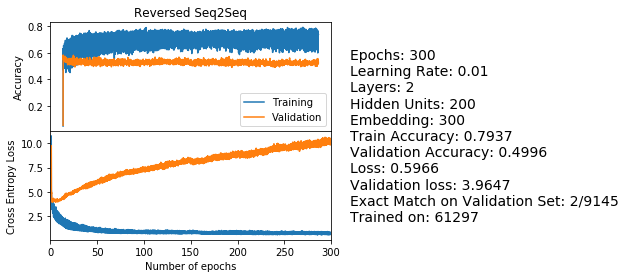}
\vspace*{0.6cm}
\includegraphics[width=0.5\textwidth]{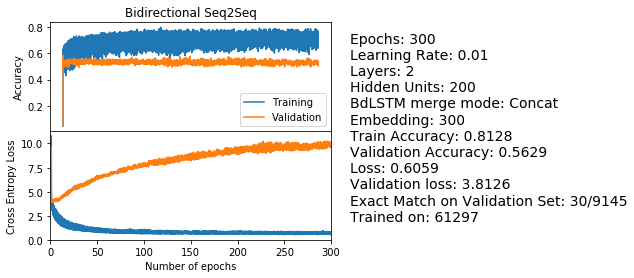}
\vspace*{0.6cm}
\includegraphics[width=0.5\textwidth]{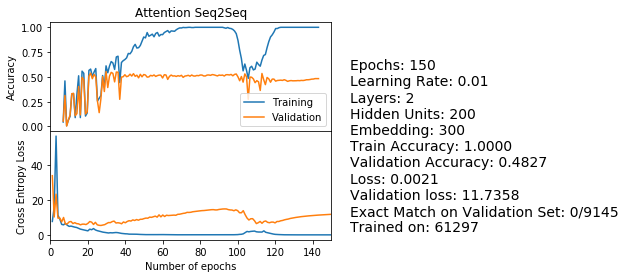}
\vspace*{0.55cm}
\includegraphics[width=0.5\textwidth]{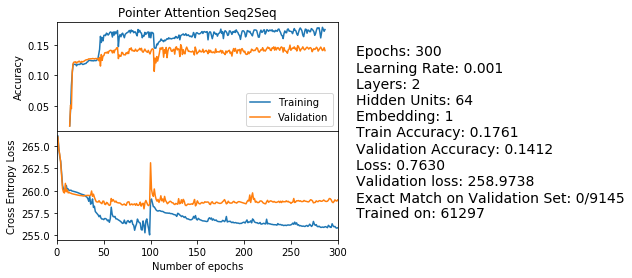}
\caption{Accuracy and Loss measurements for various models}
\end{figure}

\newpage
\subsection{Training Speed and Performance}
In terms of training speed, the Bhadanau model converged very fast, the training loss keeps dropping quickly until 0. However, this is also an indication of overfitting. Evaluating on the test set verified our insights; it performs badly on the test dataset and maps everything into a few learned output sequences. 

The vanilla and reversed seq2seq model runs the second fast but compared to the Bhadanau model it converged much slower. The dropping of training loss become slower and slower, and it will finally stopped dropping at some point. 

The pointer network runs the slowest and it took a long time to train. And because of design difficulty we have mentioned previously, it is highly sensitive to the data length and data consistency. It performs not very well on the whole training set. And after some analysis, we notice that pointer network tries to balance the output on all data point. The final result is to output the most frequent symbols appeared in all data points. This fact indicates that batch size matters. Then we tried to decrease batch size and notice increasing in training speed, but, on the other hand, we get a longer training time. In order to see the true power of this new architecture, we tried to train it on a much smaller toy set with a few data points. And we notice that even on this small dataset, it needed a long time to really figure out the position of each output word. But finally, it indeed figured out the correct pointer distribution.

Lastly, we want to notify that deep learning is compute intensive, so we kept our hyperparameters in a decent range training everything on an NVIDIA GeForce GTX 1080Ti with tensorflow 1.3, CUDA 8.0, cuDNN 6.1, python 3.6
\begin{enumerate}
\item 2 layers of LSTMs with hidden units to 200 initialized with a uniform distribution between -0.1 and 1.0 with a seed of 2.
\item Fixed number of epochs to 300 to make sure network plateau, we avoided early stopping since gradient descent guarantees local minima not global and gets stuck. We had to stop at 150 epoch for Bahdanau Attention to avoid overtraining itself.
\item Embedding size 300 which is enough to learn dense representation of our vocabulary size, also embeddings are allowed to be re-trained in each epoch.
\item Fixed batch size of 128.
\item Adam Optimizer with a Learning Rate: 0.01, reduced to 0.001 for pointer networks as it was not able to converge.
\item We are using Greedy Decoder with GreedyEmbeddingHelper which uses the \textit{argmax} of the output (treated as logits) and passes the result through an embedding layer to get the next input.
\item We are using a sequence loss which is basically a weighted cross-entropy loss for a sequence of logits.
\item Although LSTMs tend to not suffer from the vanishing gradient problem, they can have
exploding gradients, to address that we clipped our gradients at (-5.0, 5.0)
\item Different sentences have different lengths. Most sentences are short (e.g., length 20-30)
but some sentences are long (e.g., length \textgreater 100), so a minibatch of 128 randomly chosen
training sentences will have many short sentences and few long sentences, and as a result,
much of the computation in the minibatch is wasted. To address this problem, we made
sure that all sentences within a minibatch were roughly of the same length, which a 2x
speedup.
\end{enumerate}

\subsection{Sample cases}
Predictions by the models on the dev split. Q, T, and P denote the natural language question, ground truth query and it's predictions respectively. FROM table has been removed for succinctness but is predicted in entirety by our model.
\subsubsection{Vanilla seq2seq model} ~\\
\textit{After 10 epochs} \\
Q: How many schools did player number 3 play at? \\
Learns aggregation keyword placement and importance of numbers
\verbatimfont{\small}%
\begin{verbatim}
T: SELECT COUNT school/club team 
   WHERE  no. = 3
P: SELECT COUNT played 
   WHERE  played = 8
\end{verbatim}
Q: what's the division with league usl first divofion \\
Syntactical Learning: typo in the word divofion vs division
\verbatimfont{\small}%
\begin{verbatim}
T: SELECT division 
   WHERE  league = usl first division
P: SELECT division 
   WHERE  first elected = 1885
\end{verbatim}
Q: what's the u.s. open cup status for regular season of 4th, atlantic division \\
Learns sequential dependencies and year/season/2002 concepts
\verbatimfont{\small}%
\begin{verbatim}
T: SELECT u.s. open cup 
   WHERE  regular season = 4th, atlantic division
P: SELECT last division 
   WHERE  year = 2002 AND regular season = 1994/95

\end{verbatim}
\textit{After 300 epochs} \\
Q:  what was the score of the away team while playing at the arden street oval?
\verbatimfont{\small}%
\begin{verbatim}
T: SELECT away team score 
   WHERE  venue = arden street oval
P: SELECT away team score 
   WHERE  venue = arden street oval
\end{verbatim}
Q:  what was the largest crowd where the home team was fitzroy? \\
Aggregation Clause Prediction
\verbatimfont{\small}%
\begin{verbatim}
T: SELECT MAX crowd 
   WHERE  home team = fitzroy
P: SELECT MAX crowd 
   WHERE  home team = fitzroy
\end{verbatim}
Q:  which catalog was formated as a cd under the label alfa records? \\
Multiple WHERE clause prediction
\verbatimfont{\small}%
\begin{verbatim}
T: SELECT catalog 
   WHERE  label = alfa records AND format = cd
P: SELECT catalog
   WHERE  label = alfa records AND format = cd
\end{verbatim}

\subsubsection{Reversed sequence vanilla seq2seq model} ~\\
Q:    name the location attendance for january 18
\verbatimfont{\small}%
\begin{verbatim}
T: SELECT catalog 
   WHERE  label = alfa records AND format = cd
P: SELECT catalog
   WHERE  label = alfa records AND format = cd
\end{verbatim}
Q:    what was the date of the game when the away team was south melbourne?
\verbatimfont{\small}%
\begin{verbatim}
T: SELECT date  
   WHERE  away team = south melbourne
P: SELECT date  
   WHERE  away team = south melbourne
\end{verbatim}

\subsubsection{Bidirectional seq2seq model} ~\\
Q:  what scores happened on february 9?
\verbatimfont{\small}%
\begin{verbatim}
T: SELECT score 
   WHERE  date = february 9
P: SELECT score 
   WHERE  date = february 9
\end{verbatim}
Q:  when the away team is south melbourne, what's the home team score? \\
Multiple Query Resolution: The question has two question words(when, what)
\verbatimfont{\small}%
\begin{verbatim}
T: SELECT home team score 
   WHERE  away team = south melbourne
P: SELECT home team score 
   WHERE  away team = south melbourne
\end{verbatim}

\subsubsection{Bahdanau Attention seq2seq model}
\begin{verbatim}
T:  SELECT position 
    WHERE  school/club team = butler-cc-(ks) 
P:  SELECT position 
    WHERE  position = colin-actress 
    
T:  SELECT school/club team 
    WHERE  no. = 21 
P:  SELECT count # 
    WHERE  position = table 
    AND    position = 13 kong 
    
T:  SELECT count position 
    WHERE  years in toronto = 2006-07 
P:  SELECT min no. of table 
    WHERE  date = 13 
    
T:  SELECT school/club team 
    WHERE  player = amir johnson 
P:  SELECT actors name 
    WHERE  winning team = newman/haas 
           the love = 13 love = 13 kong
\end{verbatim}

The above outputs are totally irrelevant with the question, we notice that attention mechanism is sensitive to over-training and in that it learn to pay attention to highly frequent words in corpus like \textit{position} (a lot of CFL draft tables), SQL keywords (table).

These neural networks indeed learned some weak and strong concepts automagically like 'usage of aggregation operator' and 'date' or 'numbers'. This is surprising as we didn't do any semantic similarity processing or feature engineering, even the embeddings were not pre-trained.

\subsubsection{Pointer network model} ~\\
As mentioned above that our pointer network is very hard to train on a large dataset, as it tries to balance all possible correct pointers, and finally it will assign pointers to frequent common words and get into some local optimal. Since we were hacking the Tensorflow kernel code, in order to verify our design is correct, we tried to run it on a toy set with only a few data points. And we notice that it still converged very slow, but finally it can successfully figure out the correct position of each component:  

\begin{figure}[H]
   \begin{minipage}{0.475\textwidth}
     \raggedright
     \includegraphics[width=0.495\linewidth]{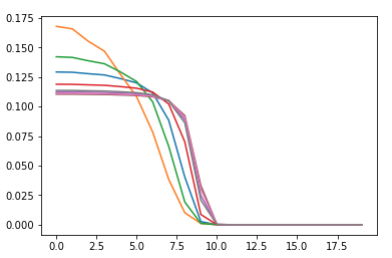}
     \includegraphics[width=0.495\linewidth]{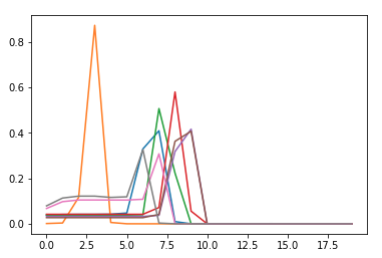}
     \includegraphics[width=0.495\linewidth]{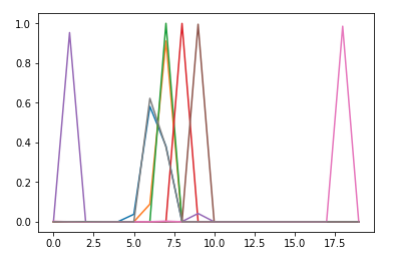}
     \includegraphics[width=0.495\linewidth]{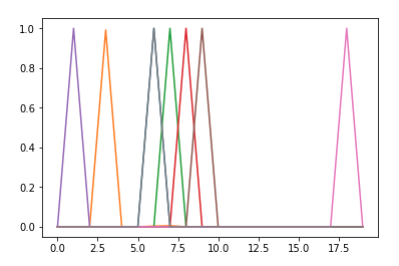}
     \caption{Position distribution of first 7 components. From left to right and top down, data sampled each 15000 steps. Though it takes a very long time to converge, but this result has at least verified the correctness of our design.}\label{Fig:Data1}
   \end{minipage}
\end{figure}

Because of using emphasizing (sequence replication), the final accuracy is not very high. However it is capable enough to cover at least one whole segment:
\begin{verbatim}
T:select max-gross-weight from where 
aircraft=robinson-r-22 select 
max-gross-weight 
from where aircraft=robinson-r-22 select 
max-gross-weight 
from where aircraft=robinson-r-22 select 
max-gross-weight 
from where aircraft=robinson-r-22 select 
max-gross-weight from

P:select max-gross-weight from where 
aircraft=robinson-r-22 select select 
select select aircraft=robinson-r-22 
select select select aircraft==select 
from select where 
aircraft=robinson-r-22 select 
max-gross-weight 
from where aircraft
\end{verbatim}
Hence we conclude that pointer network, though powerful on other tasks, might be incompatible with this sequence to sequence semantic parsing task in long sequences. Therefore we can leverage this to only predict column name from given columns and question like a question-answering system.

\subsection{LSTM memorability study}
As we have mentioned above that it is amazing that even for our vanilla seq2seq model it could already learn certain weak concepts. And we hypothesize this ability might relate with the memorability of LSTM cell itself. To verify our hypothesis, we take out the encoder only and try to intercept the output information and use it to predict the possible aggregation function used in the target SQL query only from the input natural language description. We label the target data with name of aggregation functions: NULL, MAX, MIN, COUNT, SUM, AVG and rewire the encoder to a multilayer perceptron and train the whole training dataset on this 'partial' neural network. Using embeddings naturally increased accuracy. The result, as being presented below, verified our hypothesis:
\begin{figure}[H]
\centering
\includegraphics[width=0.5\textwidth]{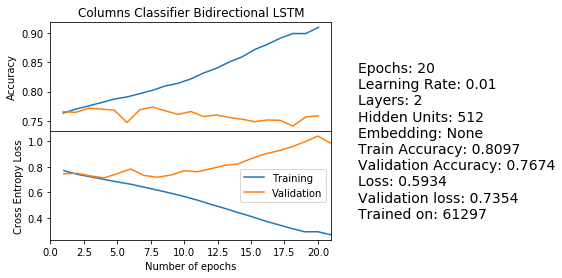}
\includegraphics[width=0.48\textwidth]{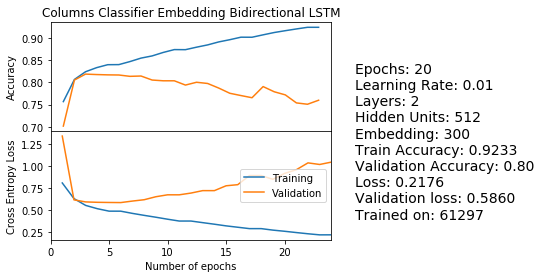}
\caption{Accuracy and loss measurement for LSTM memorability study. It only takes few epochs to achieve a very good performance. This fact also indicates that the LSTM cell indeed can capture certain concepts directly from input natural language description, which should be considered as the fundamental of all sequence to sequence models}
\end{figure}

\section{Conclusion}
We framed semantic parsing natural language questions into structured query language as a sequence to sequence neural machine translation task. We reviewed different state of art systems, techniques and evaluated their performance on WikiSQL task. We were able to parse english questions into query using just seq2seq with 56.29\% bag of words accuracy and even some exact matches. We found bidirectional seq2seq better than vanilla, reversed, and attention networks. We also observed that parsing into a SQL query is not directly a translation but a mix of both translation and word ordering problem to which we tried to apply pointer networks but could not replicate the results in paper, but we did verify that our approach and design of pointer network was working as expected. Apart from this, we also did a hypothesis testing that LSTM could encode sequences efficiently, we were able to predict aggregation operation with 80\% accuracy. Our biggest observation is that networks do not require a lot of epochs to plateau.

To evaluate our result, we wanted to use exact match as the original paper but our exact matches were negligible to correctly calibrate system performance so we recorded exact match but used positional and bag of words as accuracy measure since two queries can be equivalent even when they are not strung matching with each other. 

In future, we plan to further divide query formation tasks into dedicated networks for predicting aggregation, column using pointer networks, where clause using attention pointer networks and evaluate using Cosette \cite{cosette}, a tool to check SQL query equivalence. 

\bibliographystyle{apalike}
\bibliography{references}

\end{document}